# Effects of Nonparanormal Transform on PC and GES Search Accuracies

Joseph D. Ramsey

Department of Philosophy
Carnegie Mellon University

May 7, 2015

**Abstract**

Liu, et al., 2009 developed a transformation of a class of non-Gaussian univariate distributions into Gaussian distributions. Liu and collaborators (2012) subsequently applied the transform to search for graphical causal models for a number of empirical data sets. To our knowledge, there has been no published investigation by simulation of the conditions under which the transform aids—or harms—standard graphical model search procedures. We consider here how the transform affects the performance of two search algorithms in particular, PC (Spirtes et al., 2000; Meek 1995) and GES (Meek 1997; Chickering 2002). We find that the transform is harmless but ineffective for most cases but quite effective in very special cases for GES, namely, for moderate non-Gaussianity and moderate non-linearity. For strong non-linearity, another algorithm, PC-GES (a combination of PC with GES), is equally effective.

1. Introduction

Liu, et al. (2009) developed a transformation of a class of non-Gaussian univariate distributions into Gaussian distributions. Liu and collaborators (2012) subsequently applied the transform to search for graphical causal models for a number of empirical data sets. To our knowledge, there has been no published investigation by simulation of the conditions under which the transform aids—or harms—standard graphical model search procedures. We consider here how the transform affects the performance of two search algorithms in particular, PC (Spirtes et al., 2000; Meek 1995) and GES (Meek 1997; Chickering 2002). We find that the transform is harmless but ineffective for most cases but quite effective in very special cases for GES, namely, for moderate non-Gaussianity and moderate non-linearity. For strong non-linearity, another algorithm, PC-GES (a combination of PC with GES), is equally effective.

The transform takes a column of data whose distribution is a smooth monotone function of N(0, 1) and renders it (using order statistics) as N(m, $s^2$) where m is the mean of the data and $s^2$ is the variance of the data, possibly with standardization. There are three questions. First, how does this transform affect data that is linear and Gaussian? Second, how does the transform affect analyses for data that are within the Gaussian copula? Third, how does the transform affect data not in the Gaussian copula? Applied to an arbitrary data set, which does not necessarily fall within the theoretical purview of the transform, how bad might one expect the performance to be?

We address these questions by looking at data generated under various assumptions, running PC and GES on these data, and checking the adequacy of the resulting models. We report just four statistics. Given data set D, generated by graph G, with pattern (CP-DAG) P, yielding under the machinations of the algorithm PC or GES an output graph H, we will assess (1) the rate of false positive adjacencies (that it, the number of false adjacencies in H divided by the number of adjacencies in P); (2) the rate of adjacency recovery (that is, the number of true adjacencies in H divided by the number of adjacencies in P); (3) the arrow false positive rate (that is, the number of false arrow points in H divided by the number of arrow points in P); and (4) the arrow recovery rate (that is, the number of true arrows in H divided by the number of arrows in P). We form opinions of various searches based on these statistics.

We consider various versions of each algorithm. First of all, we consider either using the standard covariance matrix of the data as input to each algorithm, or instead using the covariance matrix of the Liu et al. transformation of the data as input to each algorithm. We denote by "PC-S" PC (Spirtes et al., 2000; Meek 1995) taking the standard covariance matrix as input and by "PC-L" PC taking the covariance matrix of the Liu et al. transformed data as input. For PC we use a significance level of 0.001. For GES (Chickering, 2002), because it is illustrative, we consider two different scores, AIC and BIC. Thus, we denote by "GES-AIC-S" GES using the AIC score taking the standard covariance matrix as input and by "GES-AIC-L" GES using the AIC score taking the covariance matrix calculated from the Liu et al. transformed data as input. Similarly for "GES-BIC-S" and "GES-BIC-L". Because it is of interest, we also show the result of running PC-S on the standard covariance matrix to get adjacencies and then running GES-BIC-S restricted to these adjacencies to get orientations. We believe from previous work that this is an amiable configuration. We denote this configuration as "PC-GES" in the tables.

We are interested in two dimensions of variation in models: distribution of disturbances, and linearity of connection functions. In all cases, we will consider directed acyclic models, since these fit under the assumptions of PC and GES. In all cases, we are interested in *structural equation models.* A simple example of a structural equation model is the following:

$X := e_X$
$Y := f(X) + e_Y$

where $X$ and $Y$ are continuous variables, f is a function, and $e_X$ and $e_Y$ are disturbances. In this case, the only sources of random variation are the disturbances, which we will take to be independent and identically distributed (i.i.d.). We may represent the above structural equation model as $e_X$->$X$->$Y$<-$e_Y$ and we may give the edge $X$->$Y$ a causal interpretation ($X$ causes $Y$) under certain assumptions.

The disturbances may be distributed in any of a number of ways, and their distribution affects the distribution of the variables $X$ and $Y$. We consider three distributions. In the default case, the disturbances are distributed Normally, which in the case of a linear model would imply that $X$ and $Y$ are also distributed Normally; we will use in particular the distribution N(0, 1) (which we will denote by "N"). Other distributions we consider will be Gamma(2, 5), and a mixture of N(-1, .5) and N(1,.5). Gamma(2, 5) is a smooth monotone transformation of N(0, 1), so it falls within the Gaussian copula and is covered by the theory given in Liu et al. We will denote Gamma(2, 5) by "NG1". The mixture of Normal distributions is univariate bimodal and cannot be obtained by a monotone function from N(0, 1). We will denote this mixture of Normal distributions by "NG2".

The other dimension of model variation we are interested in is degree of nonlinearity of the connection functions (e.g. f, above). In this case, we will consider the option that the model is linear, that is, $\Sigma_j\{a_j x_j\}$ (denoted by "L"), and two alternatives. The option that fits within the nonparanormal theory is to take $\Sigma_j\{a_i \text{abs}(x_j)|x_j|^{1.5}\}$. We will denote this by "NL1". An option that does not fit the nonparanormal theory is $\Sigma_j(a_i \sin(x_j))$; we will denote this by "NL2". In all cases, $a_j$ is drawn from U(-1, 1).

We do two analyses, one with 50 variables and 1000 cases, a case where the number of variables is far fewer than the number of nodes, and one with 500 variables and 250 cases, a case where the number of variables is considerably larger than the number of cases. The Liu et al. analysis is aimed at cases like the latter, with the goal of scaling up the number of variables that can be considered. They consider the method of estimating non-adjacencies in a causal graph by identifying zeroes in the inverse covariance matrix of the data, or in this case the inverse covariance matrix of the data transformed as per their recommendation. They apply a sparsity L1 penalty to the covariance matrices to reduce the number of false positive adjacencies in the output graph. We are interested to see whether the Liu et al. transform helps with three algorithms in particular, without a sparsity penalty-- namely, PC and GES in particular, and PC followed by GES.

The specification of the form of the AIC and BIC scores requires specification of a number of degrees of freedom. The AIC score we use is 2L - 2$k$, where L is the likelihood and $k$ is the number of degrees of freedom, and the BIC score we use is 2L - $k$ ln $n$. This is important to point out, since for the 500 variable case we will increase the penalties of these scores by a factor of 2, which we do for empirical reasons.

## 2. Results for Smaller Models

Tables 1-9 give the results of the first experiment, for data with 50 nodes and 1000 cases, for each combination of G, NG1, and NG2 with L, NL1, and NL2. In each table, row 1 shows false positive adjacency (Adj FPR) rates, row 2 adjacency recovery rates (Adj RR), row 3 false positive arrow rates (Arrow FPR), and row 4 arrow recovery rates (Arrow RR). In each table, all numbers in the same row are averages of statistics over the same 10 runs. Each run begins by selecting a new random graph with 50 nodes and 50 edges, then parameterizing it as a structural equation model with the given disturbance distributions and connection functions, then simulating 1000 cases of data, recursively, i.i.d., then conditionally doing a Liu et al. transform of the data, then forming a covariance matrix, and then running the specified algorithm with that covariance matrix as input. This is done 10 times over, errors tabulated, and then average errors calculated across runs. All simulations were carried out using the Tetrad freeware (Tetrad V, v. 5.1.0-10). We followed the implementation of huge.npn in the huge package in R (Zhao and Liu, 2012), albeit retaining means and variances of the data.

**Table 1. Accuracies for L and G for the 50 variable 1000 sample case.**

|           | PC-S | PC-L | GES-AIC-S | GES-AIC-L | GES-BIC-S | GES-BIC-L | PC-GES-S | PC-GES-L |
|-----------|------|------|-----------|-----------|-----------|-----------|----------|----------|
| Adj FPR   | 0    | 0    | 0.11      | 0.11      | 0.02      | 0.01      | 0        | 0        |
| Adj RR    | 0.78 | 0.78 | 0.89      | 0.88      | 0.81      | 0.81      | 0.78     | 0.78     |
| Arrow FPR | 0.51 | 0.51 | 0.19      | 0.18      | 0.08      | 0.08      | 0.05     | 0.05     |
| Arrow RR  | 0.63 | 0.62 | 0.71      | 0.7       | 0.59      | 0.59      | 0.55     | 0.55     |

**Table 2. Accuracies for L an NG1 for the 50 variable 1000 sample case.**

|           | PC-S | PC-L | GES-AIC-S | GES-AIC-L | GES-BIC-S | GES-BIC-L | PC-GES-S | PC-GES-L |
|-----------|------|------|-----------|-----------|-----------|-----------|----------|----------|
| Adj FPR   | 0    | 0    | 0.09      | 0.07      | 0.01      | 0.01      | 0        | 0        |
| Adj RR    | 0.76 | 0.78 | 0.87      | 0.89      | 0.8       | 0.81      | 0.76     | 0.78     |
| Arrow FPR | 0.42 | 0.47 | 0.14      | 0.21      | 0.02      | 0.03      | 0.01     | 0.03     |
| Arrow RR  | 0.61 | 0.64 | 0.74      | 0.73      | 0.61      | 0.62      | 0.59     | 0.61     |

**Table 3. Accuracies for L and NG2 for the 50 variable 1000 sample case.**

|           | PC-S | PC-L | GES-AIC-S | GES-AIC-L | GES-BIC-S | GES-BIC-L | PC-GES-S | PC-GES-L |
|-----------|------|------|-----------|-----------|-----------|-----------|----------|----------|
| Adj FPR   | 0    | 0    | 0.1       | 0.12      | 0.02      | 0.01      | 0        | 0        |
| Adj RR    | 0.77 | 0.77 | 0.88      | 0.89      | 0.8       | 0.81      | 0.77     | 0.77     |
| Arrow FPR | 0.39 | 0.37 | 0.18      | 0.22      | 0.04      | 0.04      | 0.03     | 0.03     |
| Arrow RR  | 0.67 | 0.65 | 0.78      | 0.77      | 0.67      | 0.67      | 0.64     | 0.63     |

Table 4. Accuracies for NL1 and G for the 50 variable 1000 sample case.

|  | PC-S | PC-L | GES-AIC-S | GES-AIC-L | GES-BIC-S | GES-BIC-L | PC-GES-S | PC-GES-L |
|---|---|---|---|---|---|---|---|---|
| Adj FPR | 0.02 | 0 | 0.2 | 0.11 | 0.07 | 0.01 | 0.02 | 0 |
| Adj RR | 0.74 | 0.76 | 0.9 | 0.9 | 0.83 | 0.84 | 0.74 | 0.76 |
| Arrow FPR | 0.34 | 0.31 | 0.27 | 0.14 | 0.11 | 0.02 | 0.04 | 0.01 |
| Arrow RR | 0.57 | 0.61 | 0.8 | 0.83 | 0.7 | 0.71 | 0.54 | 0.56 |

**Table 5. Accuracies for NL1 and NG1 for the 50 variable 1000 sample case.**

|  | PC-S | PC-L | GES-AIC-S | GES-AIC-L | GES-BIC-S | GES-BIC-L | PC-GES-S | PC-GES-L |
|---|---|---|---|---|---|---|---|---|
| Adj FPR | 0.23 | 0.05 | 0.94 | 0.43 | 0.67 | 0.17 | 0.22 | 0.04 |
| Adj RR | 0.53 | 0.6 | 0.74 | 0.82 | 0.71 | 0.79 | 0.52 | 0.6 |
| Arrow FPR | 0.35 | 0.15 | 1.38 | 0.62 | 1.01 | 0.24 | 0.12 | 0.05 |
| Arrow RR | 0.3 | 0.31 | 0.58 | 0.71 | 0.54 | 0.66 | 0.27 | 0.29 |

**Table 6. Accuracies for NL1 and NG2 for the 50 variable 1000 sample case.**

|  | PC-S | PC-L | GES-AIC-S | GES-AIC-L | GES-BIC-S | GES-BIC-L | PC-GES-S | PC-GES-L |
|---|---|---|---|---|---|---|---|---|
| Adj FPR | 0.02 | 0 | 0.16 | 0.12 | 0.03 | 0.02 | 0.02 | 0 |
| Adj RR | 0.74 | 0.76 | 0.89 | 0.9 | 0.83 | 0.84 | 0.74 | 0.76 |
| Arrow FPR | 0.45 | 0.43 | 0.41 | 0.35 | 0.12 | 0.1 | 0.08 | 0.08 |
| Arrow RR | 0.59 | 0.62 | 0.81 | 0.82 | 0.73 | 0.71 | 0.55 | 0.54 |

**Table 7. Accuracies for NL2 and G for the 50 variable 1000 sample case .**

|  | PC-S | PC-L | GES-AIC-S | GES-AIC-L | GES-BIC-S | GES-BIC-L | PC-GES-S | PC-GES-L |
|---|---|---|---|---|---|---|---|---|
| Adj FPR | 0 | 0 | 0.08 | 0.07 | 0 | 0 | 0 | 0 |
| Adj RR | 0.7 | 0.71 | 0.82 | 0.82 | 0.7 | 0.7 | 0.7 | 0.71 |
| Arrow FPR | 0.72 | 0.74 | 0.31 | 0.24 | 0.05 | 0.08 | 0.05 | 0.06 |
| Arrow RR | 0.54 | 0.56 | 0.56 | 0.55 | 0.32 | 0.35 | 0.43 | 0.43 |

**Table 8. Accuracies for NL2 and NG1 for the 50 variable 1000 sample case.**

|  | PC-S | PC-L | GES-AIC-S | GES-AIC-L | GES-BIC-S | GES-BIC-L | PC-GES-S | PC-GES-L |
|---|---|---|---|---|---|---|---|---|
| Adj FPR | 0 | 0 | 0.12 | 0.1 | 0 | 0 | 0 | 0 |
| Adj RR | 0 | 0 | 0.01 | 0 | 0 | 0 | 0 | 0 |
| Arrow FPR | 0 | 0 | 0 | 0 | 0 | 0 | 0 | 0 |
| Arrow RR | 0 | 0 | 0 | 0 | 0 | 0 | 0 | 0 |

**Table 9. Accuracies for NL2 and NG2 for the 50 variable 1000 sample case.**

|  | PC-S | PC-L | GES-AIC-S | GES-AIC-L | GES-BIC-S | GES-BIC-L | PC-GES-S | PC-GES-L |
|---|---|---|---|---|---|---|---|---|
| Adj FPR | 0 | 0 | 0.09 | 0.09 | 0 | 0 | 0 | 0 |
| Adj RR | 0.69 | 0.7 | 0.81 | 0.83 | 0.69 | 0.7 | 0.69 | 0.69 |
| Arrow FPR | 0.71 | 0.7 | 0.24 | 0.33 | 0.11 | 0.11 | 0.09 | 0.07 |
| Arrow RR | 0.59 | 0.58 | 0.63 | 0.63 | 0.39 | 0.37 | 0.46 | 0.45 |

## 3. Results for Larger Models

These tables suffice for the 50 variable, 1000 case scenario. We now move to the 500 variable, 250 case scenario.[1] Since the AIC GES algorithms performed universally badly for the 50 variable case, we remove them from consideration. Also, as mentioned earlier, we increase the penalty of the BIC scores by a factor of two. This results in a new set of 9 tables, Tables 10 through 18.

**Table 10. Accuracies for L and G for the 500 variable 250 sample case.**

|           | PC-S | PC-L | GES-BIC-S | GES-BIC-L | PC-GES-S | PC-GES-L |
|-----------|------|------|-----------|-----------|----------|----------|
| Adj FPR   | 0    | 0    | 0         | 0         | 0        | 0        |
| Adj RR    | 0.54 | 0.54 | 0.67      | 0.67      | 0.54     | 0.54     |
| Arrow FPR | 0.32 | 0.32 | 0.03      | 0.04      | 0        | 0        |
| Arrow RR  | 0.33 | 0.32 | 0.45      | 0.44      | 0.29     | 0.28     |

**Table 11. Accuracies for L and NG1 for the 500 variable 250 sample case**

|           | PC-S | PC-L | GES-BIC-S | GES-BIC-L | PC-GES-S | PC-GES-L |
|-----------|------|------|-----------|-----------|----------|----------|
| Adj FPR   | 0    | 0    | 0         | 0         | 0        | 0        |
| Adj RR    | 0.54 | 0.55 | 0.67      | 0.68      | 0.54     | 0.55     |
| Arrow FPR | 0.32 | 0.35 | 0.04      | 0.04      | 0.01     | 0.01     |
| Arrow RR  | 0.33 | 0.32 | 0.45      | 0.43      | 0.28     | 0.26     |

**Table 12. Accuracies for L and NG2 for the 500 variable 250 sample case**

|           | PC-S | PC-L | GES-BIC-S | GES-BIC-L | PC-GES-S | PC-GES-L |
|-----------|------|------|-----------|-----------|----------|----------|
| Adj FPR   | 0    | 0    | 0         | 0         | 0        | 0        |
| Adj RR    | 0.54 | 0.55 | 0.67      | 0.68      | 0.54     | 0.55     |
| Arrow FPR | 0.31 | 0.35 | 0.03      | 0.03      | 0        | 0        |
| Arrow RR  | 0.32 | 0.32 | 0.44      | 0.43      | 0.28     | 0.27     |

**Table 13. Accuracies for NL1 and G for the 500 variable 250 sample case**

|           | PC-S | PC-L | GES-BIC-S | GES-BIC-L | PC-GES-S | PC-GES-L |
|-----------|------|------|-----------|-----------|----------|----------|
| Adj FPR   | 0.01 | 0    | 0.03      | 0.01      | 0.01     | 0        |
| Adj RR    | 0.53 | 0.53 | 0.71      | 0.72      | 0.52     | 0.53     |
| Arrow FPR | 0.2  | 0.2  | 0.07      | 0.03      | 0        | 0        |
| Arrow RR  | 0.25 | 0.26 | 0.54      | 0.54      | 0.22     | 0.23     |

---

[1] We limit ourselves to 500 variables because of the limitations of the nonlinear, non-Gaussian simulator that we used.

**Table 14. Accuracies for NL1 and NG1 for the 500 variable 250 sample case**

|          | PC-S | PC-L | GES-BIC-S | GES-BIC-L | PC-GES-S | PC-GES-L |
|----------|------|------|-----------|-----------|----------|----------|
| Adj FPR  | 0.14 | 0.02 | 0.47      | 0.1       | 0.14     | 0.02     |
| Adj RR   | 0.41 | 0.46 | 0.63      | 0.74      | 0.41     | 0.46     |
| Arrow FPR| 0.16 | 0.06 | 0.71      | 0.17      | 0.04     | 0        |
| Arrow RR | 0.12 | 0.13 | 0.44      | 0.61      | 0.1      | 0.12     |

**Table 15. Accuracies for NL1 and NG2 for the 500 variable 250 sample case**

|          | PC-S | PC-L | GES-BIC-S | GES-BIC-L | PC-GES-S | PC-GES-L |
|----------|------|------|-----------|-----------|----------|----------|
| Adj FPR  | 0.01 | 0    | 0.02      | 0.01      | 0.01     | 0        |
| Adj RR   | 0.53 | 0.55 | 0.72      | 0.74      | 0.53     | 0.55     |
| Arrow FPR| 0.2  | 0.24 | 0.04      | 0.02      | 0        | 0        |
| Arrow RR | 0.28 | 0.28 | 0.54      | 0.56      | 0.24     | 0.25     |

**Table 16. Accuracies for NL2 and G for the 500 variable 250 sample case**

|          | PC-S | PC-L | GES-BIC-S | GES-BIC-L | PC-GES-S | PC-GES-L |
|----------|------|------|-----------|-----------|----------|----------|
| Adj FPR  | 0    | 0    | 0         | 0         | 0        | 0        |
| Adj RR   | 0.35 | 0.37 | 0.44      | 0.45      | 0.35     | 0.37     |
| Arrow FPR| 0.21 | 0.24 | 0.02      | 0.02      | 0        | 0        |
| Arrow RR | 0.16 | 0.18 | 0.07      | 0.08      | 0.08     | 0.1      |

**Table 17. Accuracies for NL2 and NG1 for the 500 variable 250 sample case**

|          | PC-S | PC-L | GES-BIC-S | GES-BIC-L | PC-GES-S | PC-GES-L |
|----------|------|------|-----------|-----------|----------|----------|
| Adj FPR  | 0    | 0    | 0         | 0         | 0        | 0        |
| Adj RR   | 0    | 0    | 0         | 0         | 0        | 0        |
| Arrow FPR| 0    | 0    | 0         | 0         | 0        | 0        |
| Arrow RR | 0    | 0    | 0         | 0         | 0        | 0        |

**Table 18. Accuracies for NL2 and NG2 for the 500 variable 250 sample case**

|          | PC-S | PC-L | GES-BIC-S | GES-BIC-L | PC-GES-S | PC-GES-L |
|----------|------|------|-----------|-----------|----------|----------|
| Adj FPR  | 0    | 0    | 0         | 0         | 0        | 0        |
| Adj RR   | 0.41 | 0.4  | 0.5       | 0.5       | 0.41     | 0.4      |
| Arrow FPR| 0.31 | 0.3  | 0.03      | 0.02      | 0.01     | 0.01     |
| Arrow RR | 0.22 | 0.19 | 0.12      | 0.08      | 0.13     | 0.1      |

4. Discussion.

Tables 1-18 provide enough information to give a preliminary answer to our earlier questions.

*(a) How does the Liu et al. transform affect data that is linear and Gaussian?*

The general answer is, not much, which is not surprising, since the transform should take the data to itself (perhaps standardized). The primary difference is between the methods themselves. This is true both for the 50 variable case as well as the 500 variable case.

*(b) How does the transform affect analyses for data most likely produce nonparanormal distributions of the variables?*

This includes the following combinations: L/NG1, NL1/G, NL1/NG1. To regale the cases for 50 variables, we have the following.

For L and NG1, for the 50 variable case (Table 2), the Liu et al. transform helps to lower the false positive ratio, but does not affect PC-GES. For the 500 variable case (Table 11), the Liu et al. transform makes no difference, and GES-BIC has the best statistics.

For NL1 and G for the 50 variable case (Table 4), the Liu et al. transform again helps GES-BIC by reducing false positives. (GES-AIC is helped too but shows worse performance.) PC-GES is not particularly helped. Overall GES-BIC-L has the best performance. For the 500 variable case (Table 13), for GES-BIC, false positives are helped some by the Liu et al. transform.

For NL1 and NG1 (Table 5), the Liu et al. transform helps all four algorithms noticeably, rendering what are otherwise quite bad false positive statistics manageable and boosting recall statistics. The best algorithm depends on whether one wants low false positives (PC-GES) or high recall (GES-BIC), with the Liu et al. transform. For the 500 variable case (Table 14), the Liu et al. transform similarly improves model statistics considerably.

*(3) How does the Liu et al. transform affect data where the variables are not likely to be nonparanormal?*

This involves potentially any rows in the table involving NG2 or NL2. We regale the cases for 50 variables as follows:

For L and NG2 (Table 3), the Liu et al. transform does not affect accuracy. By a thin margin, GES is the best algorithm (with the Liu et al. transform). A similar comment holds for the 500 variable case (Table 11).

For NL1 and NG2, for the 50 varible case (Table 6), the Liu et al. transform does not have much of an effect. The best algorithm is GES-BIC. The 500 variable case (Table 15) shows a similar result.

For NL2 and G (Table 7), PC-GES is the clear winner, by a small margin. The Liu et al transform does not have an effect. The 500 variable case (Table 16) shows the same result.

For NL2 and NG1 (Table 8, Table 17), all options are bad.

For NL2 and NG2, the Liu et al. transform has no effect, except to make the false positive rate for GES-AIC worse. For the 50 variable case (Table 9), PC-GES is the best option on recall. For the 500 variable case, GES-BIC and PC-GES are both good options.

Across all of these tables except for GES-BIC is a good option. Using the Liu et al. transform is in no case especially harmful and in many cases does not affect accuracy at all. Since the Liu et al. transform with GES is especially helpful for the case of mild non-Gaussianity and mild nonlinearity, it can be recommended in situations where those conditions might obtain. For more severe nonlinearities, either GES or PC-GES appears to be the algorithm of choice. These observations hold for the 50 variable case as well as the 500 variable case.

It is interesting to note how well GES performs across so many of the conditions surveyed; this stands in need of an explanation, which we do not provide here.

**5. Conclusion**

The Liu transform is ingenious and may have useful applications in estimation and prediction. For the, admittedly limited in graph complexity, sample of conditions we have examined here, however, it does no harm and is of positive aid especially for the moderate nonlinear, non-Gaussian case.